# WAIR-D: Wireless AI Research Dataset

Yourui Huangfu*, Jian Wang*, Shengchen Dai*, Rong Li*, Jun Wang*, Chongwen Huang†, Zhaoyang Zhang†

*Wireless Technology Laboratory, Huawei Technologies Co., Ltd., Hangzhou, China*

{huangfuyourui, wangjian23, daishengchen, lirongone.li, justin.wangjun}@huawei.com,*

*Zhejiang University, Hangzhou, China†*

{chongwenhuang, ning_ming}@zju.edu.cn†

*Abstract* — It is a common sense that datasets with high-quality data samples play an important role in artificial intelligence (AI), machine learning (ML) and related studies. However, although AI/ML has been introduced in wireless researches long time ago, few datasets are commonly used in the research community. Without a common dataset, AI-based methods proposed for wireless systems are hard to compare with both the traditional baselines and even each other. The existing wireless AI researches usually rely on datasets generated based on statistical models or ray-tracing simulations with limited environments. The statistical data hinder the trained AI models from further fine-tuning for a specific scenario, and ray-tracing data with limited environments lower down the generalization capability of the trained AI models. In this paper, we present the Wireless AI Research Dataset (WAIR-D)[1], which consists of two scenarios. Scenario 1 contains 10,000 environments with sparsely dropped user equipments (UEs), and Scenario 2 contains 100 environments with densely dropped UEs. The environments are randomly picked up from more than 40 cities in the real world map. The large volume of the data guarantees that the trained AI models enjoy good generalization capability, while fine-tuning can be easily carried out on a specific chosen environment. Moreover, both the wireless channels and the corresponding environmental information are provided in WAIR-D, so that extra-information-aided communication mechanism can be designed and evaluated. WAIR-D provides the researchers benchmarks to compare their different designs or reproduce results of others. In this paper, we show the detailed construction of this dataset and examples of using it.

## I. INTRODUCTION

In the fundamental of artificial intelligence (AI) and machine learning (ML) related studies, datasets with high-quality data samples have been playing one of the most important roles in AI model design, training, evaluation, benchmarking and deployment. Referring to all the applications where AI/ML-based methods show superior performance, e.g. natural language processing (NLP) and computer vision (CV), we can always find the participation of commonly used high-quality datasets. These datasets serve as the fuel of the whole eco-system, hence should be constructed firstly whenever AI/ML is adopted in a new research domain such as wireless communications.

Bringing AI/ML into wireless communication systems has drawn a lot of attentions, and we can find corresponding AI/ML-based research works for the physical layer (PHY), MAC layer and other higher layers in wireless system. For example, in PHY, neural networks (NNs) have been used to replace the end-to-end transceiver [1, 2] and some particular modules such as channel modelling [3], channel estimation [4], MIMO detector [5], and so on. AI/ML-based methods also show good performance on scheduling and radio resource management (RRM) tasks [6, 7, 8], where reinforcement learning (RL) is usually considered. Moreover, viewing the wireless systems as ones with various distributed nodes such as mobile phones, laptops, sensors, cars and drones equipped with radio transceivers, it also serves as a perfect application scenario for distributed learning frameworks [9].

Despite there is a lot of work focusing on applying AI/ML in wireless systems, most of the studies are performed in monotonous propagation environments. Compared to the various commonly accepted datasets in domains such as NLP, CV and medicine, only a few datasets are available for wireless system study, such as DeepMIMO [10], Raymobtime [11] and RadioML [12]. However, the diversity of propagation environments in these existing datasets is not so good, hence they are unsuitable to be used in the cases where the good generalization capability of AI/ML models is expected.

In this paper, we introduce the Wireless AI Research Dataset (WAIR-D), which aims to help researchers to study and evaluate AI/ML algorithms for wireless systems. The most important features of this dataset include:

- Radio propagation paths are generated through a ray-tracing simulator with given environment settings, hence both channel data and the corresponding environmental information can be obtained from WAIR-D. The researchers are able to rely on WAIR-D for both environment-irrelevant and environment-relevant tasks.
- Channels can be generated with specific wireless system parameters, which can be set by the users of WAIR-D. Five possible carrier frequencies (2.6GHz, 6GHz, 28GHz, 60GHz and 100GHz) are supported, which makes WAIR-D suitable for the use in both sub-6Hz and millimeter wave applications. Other parameters such as the system bandwidth, the numerology, and the number of antennas can be chosen according to the application of interest.
- Two general scenarios are considered in WAIR-D, i.e., Scenario 1 with 5 base stations (BSs) and 30 sparsely dropped user equipments (UEs) in each environment, and Scenario 2 with 1 BS and 10,000 densely dropped UEs in each environment. AI models trained with data from one scenario can be fine-tuned with data from the other scenario. Transfer learning can be tested relying on the good diversity of WAIR-D.
- Real world maps are used to provide building layout information, which makes WAIR-D closer to the reality. Moreover, a large volume of environments, i.e., 10,000 for Scenario 1 and 100 for Scenario 2, are considered, hence the generalization capability of AI algorithms is easily to be evaluated based on WAIR-D.

The rest of the paper is organized as follows: we provide a detailed description for WAIR-D in Section II. In Section III, two examples of using the proposed WAIR-D, i.e., AI-based wireless environment reconstruction and spatial beam prediction, are provided. Finally, we draw some conclusions in Section IV.

---

1. The latest version of WAIR-D can be found at: https://www.mobileai-dataset.com/html/default/yingwen/DateSet/1590994253188792322.html?index=1

## II. A DETAILED DESCRIPTION FOR WAIR-D

WAIR-D aims to provide a wireless dataset that has various environments which are close to the real ones. To achieve this goal, we make use of a 3D ray-tracing simulator called PyLayers [13]. It takes environmental information, such as a map with the locations of the BSs and UEs, as input and outputs the corresponding radio propagation paths for each radio link between each BS and UE. To make the simulation closer to the reality, the public map service, OpenStreetMap (OSM) [14], is adopted. The information about the layout of real buildings and directions of real streets from OSM can help with improving the authenticity and diversity of the data.

In this section, we provide a detailed description for WAIR-D, which includes the construction procedure, the scenarios considered, the dataset structure and the method of user-specific channel data generation.

### A. Construction procedure

To generate data for a specific environment, we first let the ray-tracing simulator randomly visit a location from more than 40 biggest cities around the world in OSM. With the chosen location, the simulator reads the descriptions of the buildings and streets from the OSM service. Depending on the randomly picked locations, from 2 to dozens of buildings may be included in each environment. As region size of each environment is unique, the region size value is provided as a part of the environmental information in the dataset.

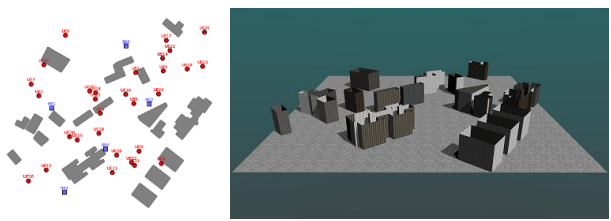

Figure 1 An example environment with buildings and BS(blue)/UE(red) positions in top view (left), the same buildings in bird-eye view (right)

BSs and UEs are then randomly dropped in each environment, where an example is illustrated in Figure 1. The locations of BSs and UEs are recorded is the dataset. Between each BS and UE, there is one radio link, which should contain at least one radio propagation path. Otherwise, the UE would be dropped again to get a new position instead. The height of BS is fixed to 6 meters, while UE is 1.5 meters high. Both of them are equipped with Omni-direction antennas.

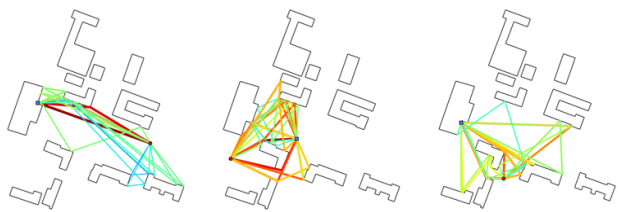

Figure 2 Visualization of simulated radio propagation paths for three radio links in an environment (darker and thicker lines indicate stronger radio propagation paths)

The simulated radio propagation paths for 3 radio links in an environment are shown in Figure 2 as an example. From the visualization of radio propagation paths, we can see that these paths are formed through three possible ways of radio propagation, i.e., direct propagation, reflection and diffraction. Both line of sight (LOS) paths and NLOS path may exist for one radio link. Darker and thicker lines indicate stronger paths, where LOS path, if any, of course is the strongest one.

### B. Scenarios in WARI-D

WAIR-D consists of two scenarios, i.e., Scenario 1 with sparsely dropped UEs and Scenario 2 with densely dropped UEs. These two scenarios are provided to guarantee the diversity and facilitate the evaluation of generalization for different wireless AI algorithms.

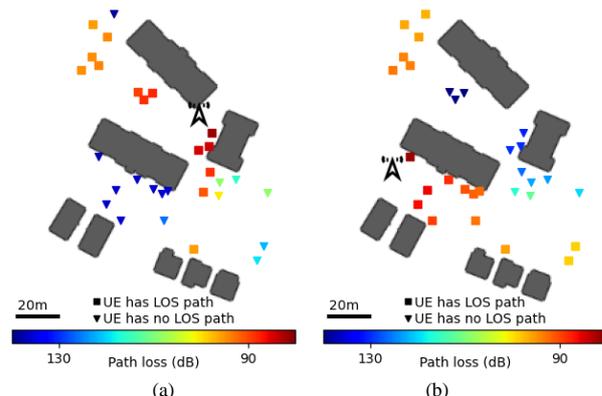

Figure 3. In the first environment in Scenario 1, (a) shows all the 30 UEs with the first BS, (b) shows all the 30 UEs with the second BS

*1) Scenario 1 (diverse environments with the sparsely dropped UEs):* In Scenario 1, 10,000 different environments are considered, each of which contains 5 BSs and 30 sparsely dropped UEs. Therefore, there are 150 radio links in each environment in total. Different radio links may contain different numbers of radio propagation paths. The LOS or non-LOS (NLOS) label of each radio propagation path can be decided based on the provided data, which is quite useful for beam-related tasks. An example of the first environment in Scenario 1 is shown in Figure 3, where the 8 dark gray blocks are buildings. Figure 3(a) shows all 30 UEs and the first BS, while Figure 3(b) shows the same UEs with the second BS. UEs with LOS path to the BS are marked by squares, and those without LOS path are marked by triangles. The path loss value (at 28GHz) for each radio link is calculated and mapped to different colors. Another important environmental information, i.e. the scale, is also calculated according to the region size of this environment and shown in the figures.

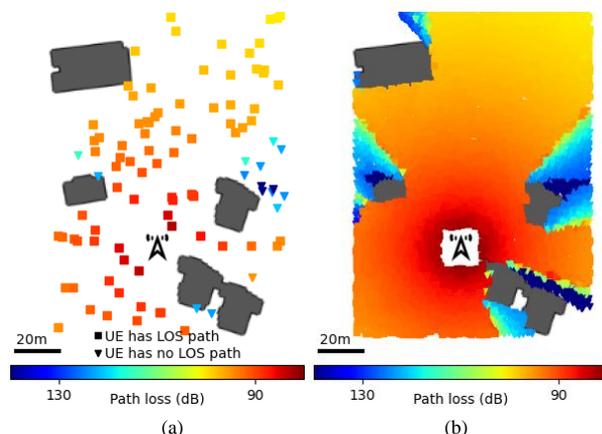

Figure 4. An example environment in Scenario 2, (a) shows 100 UEs among the totally 10,000 UEs, (b) shows all 10,000 UEs

*2) Scenario 2 (specific environments with the densely dropped UEs):* In Scenario 2, there are 100 environments, which are picked up from the 10,000 environments in Scenario 1. In each environment, only one BS is randomly dropped together with 10,000 densely dropped UEs. The resulting 10,000 radio links in each environment can be used to train a dedicated model for the corresponding environment. Obviously, the totally 1 million radio links can also be used together if needed. In Figure 4(a), 100 UEs among all the 10,000 UEs and the BS are shown. For all 10,000 UEs, since the UE positions are quite dense, we only show the path loss value (at 28GHz) through color mapping in Figure 4(b).

The data of totally 1.5 million radio links, i.e., 10,000 environments with 150 radio links in each, makes Scenario 1 more suitable to be used to train AI models with good generalization capability. As the AI models have "seen" more and more environments, they are expected to work in other unseen environments better and better. If an AI model is supposed to be trained under a certain environment, Scenario 2 is more helpful.

C. Dataset structure

As shown in Figure 5, WAIR-D mainly contains three parts, i.e., the raw data, an example task, and scripts for user-specific data generation.

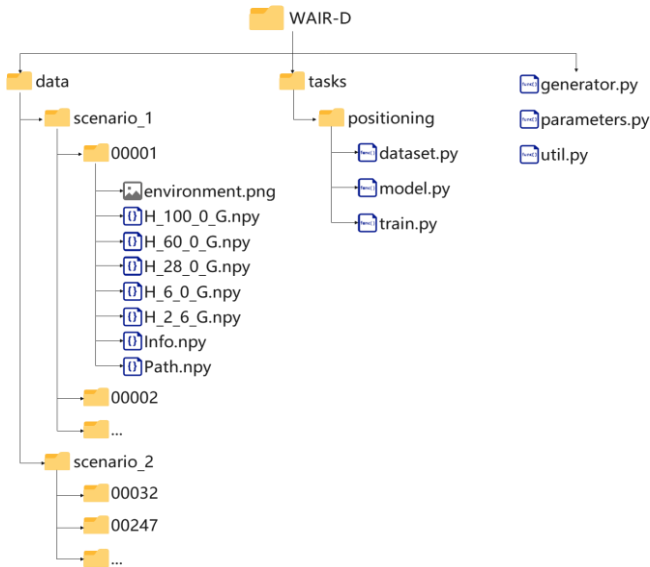

Figure 5 The file structure of WAIR-D

The raw data for each environment is provided within a separate folder. Taking Scenario 1 as an example, for the first environment, all the raw data is stored in the folder named "data/scenario_1/00001". The map for each environment is stored in the image file named "environment.png". The environmental information, i.e., the positions of the BSs and UEs, as well as the region size of this environment, is in the file named "Info.npy". The file named "Path.npy" contains the path delay, angle of arrival (AoA), angle of departure (AoD) for each radio propagation path of all the radio links. The path responses for all the radio propagation paths are simulated under five possible carrier frequencies (cf = {2.6GHz, 6GHz, 28GHz, 60GHz, 100GHz}), and the simulated data is recorded in the files named "H_cf_G.npy" with "cf" indicating the carrier frequency, respectively.

The positioning task is provided as an example of how to use WAIR-D. The files named "dataset.py", "model.py" and "train.py" in the folder named "tasks/positioning" give sample codes of data preparation, AI/ML model construction and training.

The channel data with user-specific wireless system settings can be generated based on the raw data in WAIR-D. The related scripts, i.e., "generator.py", "parameters.py" and "util.py", are also provided, the method of generating wireless channel will be introduced in Section II-D.

D. User-specific Channel Data Generation

As described in Section II-C, from the raw data in WAIR-D, we can load the path delays, AoAs, AoDs and path responses for all the radio propagation paths. Based on this path information, we can calculate channel frequency response (CFR) for each radio link through

$$H_f = \sum_r p_{r,cf} \times e^{-2\pi j \tau_r f}$$

where, $p_{r,cf}$ and $\tau_r$ are the path response and path delay for path $r$ of the considered radio link respectively, $cf$ is the carrier frequency, and $f$ is a vector of frequency points which can be set by the users. For example, considering the Orthogonal Frequency Division Multiplexing (OFDM) system, given the bandwidth, subcarrier spacing and carrier frequency, the frequency points of all the subcarriers can be calculated and concatenated into the vector $f$.

Through the above-mentioned equation, the frequency dimension is added while the path dimension is eliminated. Additionally, uniform planar array (UPA) or other antenna response matrices can be calculated from the AoAs and AoDs and multiplied on the channel matrix to introduce the antenna dimension.

III. EXAMPLE APPLICATIONS OF WAIR-D

In this section, environment reconstruction task and spatial beam prediction task are introduced as application examples of WAIR-D. As shown in Table 1, the user-specific system parameters, such as carrier frequency, bandwidth and antennas, are used for generating the channel matrices.

TABLE 1. Parameters for tasks

| Parameters for | Environment reconstruction | Spatial beam prediction |
|---|---|---|
| Carrier Frequency | 6GHz | 28GHz |
| Bandwidth | 31.25MHz | 46.08MHz |
| Num. of OFDM subcarriers | 64 | 384 |
| Antennas | 16T4R | 64T1R |
| Antenna spacing | Half wavelength | Half wavelength |
| Links used for reconstruction | 150 | ~ |

A. Environment reconstruction

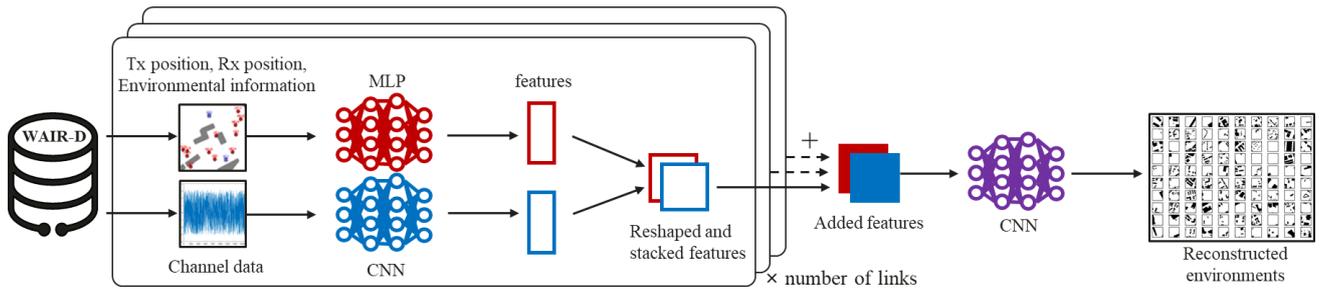

Figure 6 The NN structure for environment reconstruction task

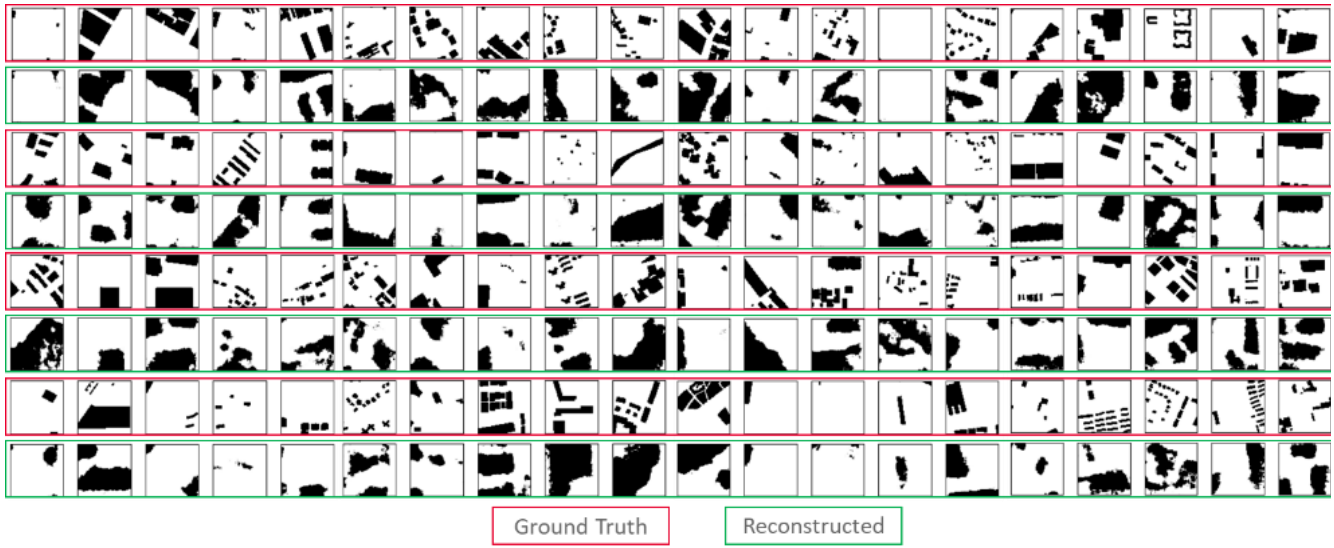

Figure 7 Comparison of ground-truth and reconstructed environments

To construct WAIR-D, the radio links are simulated with a given environment. While, it is suspected that whether the inverse problem, i.e. reconstruction the environment from its corresponding ratio links, is also solvable. In this subsection, we show some promising results on this task with the help of data from WAIR-D Scenario 1.

In a real situation, the number of available radio links for the environment reconstruction task may be time-varying. Therefore, to support arbitrary number of radio links as input, we propose a neural network (NN) structure as shown in Figure 6. For each radio link, a multilayer perceptron (MLP) is used to process the transceiver positions, as well as the scale of environment. Meanwhile, a convolutional neural network (CNN) is used to extract features from the channel data. After processing all available links repeatedly with these two NNs, outputs generated from all the radio links are then added together and passed through another CNN. The output of the proposed NN is the reconstructed environment, which is compared to the ground truth from the dataset for loss computation. Supervised learning method is used to train the NN.

After training with all 150 radio links as inputs for each environment to be reconstructed in the training set, we tested the trained model on the testing set. As shown in Figure 7, it is shown that the reconstructed environment have the similar positions and layouts with the real environment, which means AI/ML can help with using this inverse problem. Meanwhile, It is found that, some regions behind buildings are hard to be reconstructed. This is because there may be no radio propagation paths related to these regions, which makes it impossible to reconstruct. To further increase the reconstruction accuracy, researchers can develop better solutions by optimizing NN model design, or increasing the angular and delay resolution through increasing antenna numbers and bandwidth.

B. Spatial beam prediction

Model transfer and generalization are important issues that need to be evaluated before any AI/ML algorithm being adopted in real systems. In this subsection, we take spatial beam prediction task as an example to show how to study these problems with WAIR-D.

As the number of beams increases in massive MIMO systems, measuring the signal quality of all the beams is resource-consuming. Instead, the spatial beam prediction mechanism is proposed, where a sparse measurement can be conducted on a few beams, and the signal quality of other unmeasured beams can be predicted based on the results from this sparse measurement. It is typically a two-stage procedure. In stage one, a part of all the beams, e.g., 16 out of 64 beams, are measured. Then, in stage two, with the signal quality of other unmeasured beams predicted, several beams are chosen as the candidates for the following transmission.

In this trial, the number of total beams is 64, and only 16 of them are measured in stage one. In stage two, $K \in \{1,2,3,5,7\}$ candidate beams are chosen. If the strongest beam is indeed in the chosen candidates, we say the prediction is correct, otherwise false. The average accuracy is calculated based on the inference results on the testing set.

We firstly pre-train a model based on 1,000 environments from WAIR-D Scenario 1, and then try to transfer the pre-trained model to a specific environment from WAIR-D Scenario 2. The results are shown in Figure 8. With the fine-tuning step being zero, it means directly using the pre-trained model in the new environment. As can be seen from Figure 8, with a few steps of fine-tuning, the prediction accuracy of the fine-tuned model can be improved dramatically, which means transfer learning with few-shot fine-tuning is helpful.

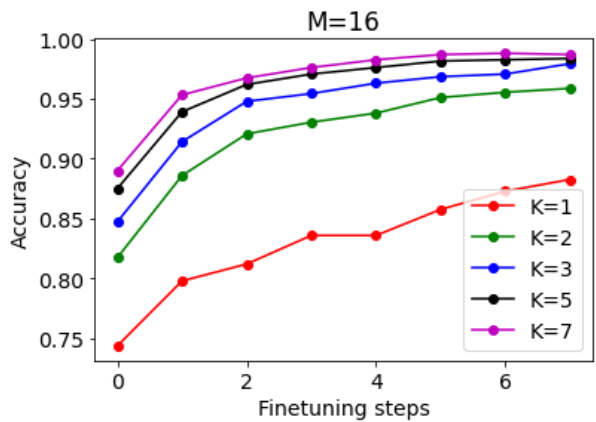

Figure 8 Performance of model transfer in the spatial beam prediction task

## IV. CONCLUSIONS

In this paper, we presented the Wireless AI Research Dataset (WAIR-D), which is a wireless AI research dataset generated with various environments. We briefly explained the dataset construction and two application examples of WAIR-D. In future, we will show more applications based on this dataset, and more scenarios and data will be added into it.

REFERENCES

[1] Hu, B., Wang, J., Xu, C., Zhang, G., & Li, R. (2021, September). A Kalman-based Autoencoder Framework for End-to-End Communication Systems. In 2021 IEEE 32nd Annual International Symposium on Personal, Indoor and Mobile Radio Communications (PIMRC) (pp. 1-6). IEEE.

[2] Ge, Y., Shi, W., Wang, J., Li, R., & Tong, W. (2022, July). Joint Message Passing and Auto-Encoder for Deep Learning. In Proceedings of the First International Workshop on Artificial Intelligence in Beyond 5G and 6G Wireless Networks (AI6G 2022).

[3] Huangfu, Y., Wang, J., Xu, C., Li, R., Ge, Y., Wang, X., ... & Wang, J. (2019, December). Realistic channel models pre-training. In 2019 IEEE Globecom Workshops (GC Wkshps) (pp. 1-6). IEEE.

[4] Huangfu, Y., Wang, J., Li, R., Xu, C., Wang, X., Zhang, H., & Wang, J. (2019, September). Predicting the mumble of wireless channel with sequence-to-sequence models. In 2019 IEEE 30th Annual International Symposium on Personal, Indoor and Mobile Radio Communications (PIMRC) (pp. 1-7). IEEE.

[5] Wang, J., Li, R., Wang, J., Ge, Y. Q., Zhang, Q. F., & Shi, W. X. (2020). Artificial intelligence and wireless communications. Frontiers of Information Technology & Electronic Engineering, 21(10), 1413-1425.

[6] Wang, J., Xu, C., Huangfu, Y., Li, R., Ge, Y., & Wang, J. (2019, October). Deep reinforcement learning for scheduling in cellular networks. In 2019 11th International Conference on Wireless Communications and Signal Processing (WCSP) (pp. 1-6). IEEE.

[7] Xu, C., Wang, J., Yu, T., Kong, C., Huangfu, Y., Li, R., ... & Wang, J. (2020, May). Buffer-aware wireless scheduling based on deep reinforcement learning. In 2020 IEEE Wireless Communications and Networking Conference (WCNC) (pp. 1-6). IEEE.

[8] Wang, J., Xu, C., Li, R., Ge, Y., & Wang, J. (2021, September). Smart scheduling based on deep reinforcement learning for cellular networks. In 2021 IEEE 32nd Annual International Symposium on Personal, Indoor and Mobile Radio Communications (PIMRC) (pp. 1-6). IEEE.

[9] Wang, J., Huangfu, Y., Li, R., Ge, Y., & Wang, J. (2021, December). Distributed Learning for Time-varying Networks: A Scalable Design. In 2021 IEEE Globecom Workshops (GC Wkshps) (pp. 1-5). IEEE.

[10] Alkhateeb, A. (2019). DeepMIMO: A generic deep learning dataset for millimeter wave and massive MIMO applications. arXiv preprint arXiv:1902.06435.

[11] Klautau, A., Batista, P., González-Prelcic, N., Wang, Y., & Heath, R. W. (2018, February). 5G MIMO data for machine learning: Application to beam-selection using deep learning. In 2018 Information Theory and Applications Workshop (ITA) (pp. 1-9). IEEE.

[12] O'Shea, T. J., Roy, T., & Clancy, T. C. (2018). Over-the-air deep learning based radio signal classification. IEEE Journal of Selected Topics in Signal Processing, 12(1), 168-179.

[13] Amiot, N., Laaraiedh, M., & Uguen, B. (2013, June). Pylayers: An open source dynamic simulator for indoor propagation and localization. In 2013 IEEE International Conference on Communications Workshops (ICC) (pp. 84-88). IEEE.

[14] Haklay, M., & Weber, P. (2008). https://www.openstreetmap.org/. Openstreetmap: User-generated street maps. IEEE Pervasive computing, 7(4), 12-18.